\documentclass{article}
\usepackage{ijcai22}
\usepackage{algorithm2e}
\usepackage{times}
\usepackage{soul}
\usepackage{url}
\usepackage[hidelinks]{hyperref}
\usepackage[utf8]{inputenc}
\usepackage[small]{caption}
\usepackage{graphicx}
\usepackage{amsmath}
\usepackage{booktabs}
\usepackage{natbib}
\usepackage{amsfonts}

\title{Alterfactual Explanations - \\The Relevance of Irrelevance for Explaining AI Systems}

\author{
Silvan Mertes$^1$\footnote{Contact Author}\and
Christina Karle\and
Tobias Huber$^{1}$\and
Katharina Weitz$^{1}$\and \\
Ruben Schlagowski$^{1}$\and
Elisabeth André$^1$\\
\affiliations
$^1$Chair for Human-Centered Artificial Intelligence, Augsburg University\\
\emails
\{first, second\}@uni-a.de
}

\begin{document}

\maketitle

\begin{abstract}
Explanation mechanisms from the field of Counterfactual Thinking are a widely-used paradigm for Explainable Artificial Intelligence (XAI), as they follow a natural way of reasoning that humans are familiar with. However, all common approaches from this field are based on communicating information about features or characteristics that are especially important for an AI's decision. We argue that in order to fully understand a decision, not only knowledge about relevant features is needed, but that the awareness of irrelevant information also highly contributes to the creation of a user's mental model of an AI system. Therefore, we introduce a new way of explaining AI systems. Our approach, which we call Alterfactual Explanations, is based on showing an alternative reality where irrelevant features of an AI's input are altered. By doing so, the user directly sees which characteristics of the input data can change arbitrarily without influencing the AI's decision. We evaluate our approach in an extensive user study, revealing that it is able to significantly contribute to the participants' understanding of an AI. 
We show that alterfactual explanations are suited to convey an understanding of different aspects of the AI's reasoning than established counterfactual explanation methods.



\end{abstract}

\section{Introduction}

With the steady advance of Artificial Intelligence (AI), and the resulting introduction of AI-based applications into everyday life, more and more people are being directly confronted with decisions made by AI algorithms \citep{stone2016one}. As the field of AI advances, so does the need to make such decisions explainable and transparent. The development and evaluation of \emph{Explainable AI} (XAI) methods is important not only to provide end users with explanations that increase acceptance and trust in AI-based methods, but also to empower researchers and developers with insights to improve their algorithms.

The need for XAI methods has prompted the research community to develop a bewildering number of different approaches to unraveling the black boxes of AI models. A considerable part of these approaches is based on telling the user of the XAI system in various ways \emph{which} features of the input data are important for a decision (often called \emph{Feature Attribution}) \citep{arrieta2020explainable}. Other methods, which are closer to human habits of explanation, are based on the paradigm of \emph{Counterfactual Thinking} \citep{miller2019explanation}. 
Procedures that follow this guiding principle try answering the question of \emph{What if...?} by showing an alternative reality and the corresponding decision of the AI.
Here, in contrast to feature attribution mechanisms, not only the importance of the various features is emphasized. Rather, it is conveyed, even if only indirectly, \emph{why} features are relevant.
\begin{figure}
 \includegraphics[width=0.48\textwidth]{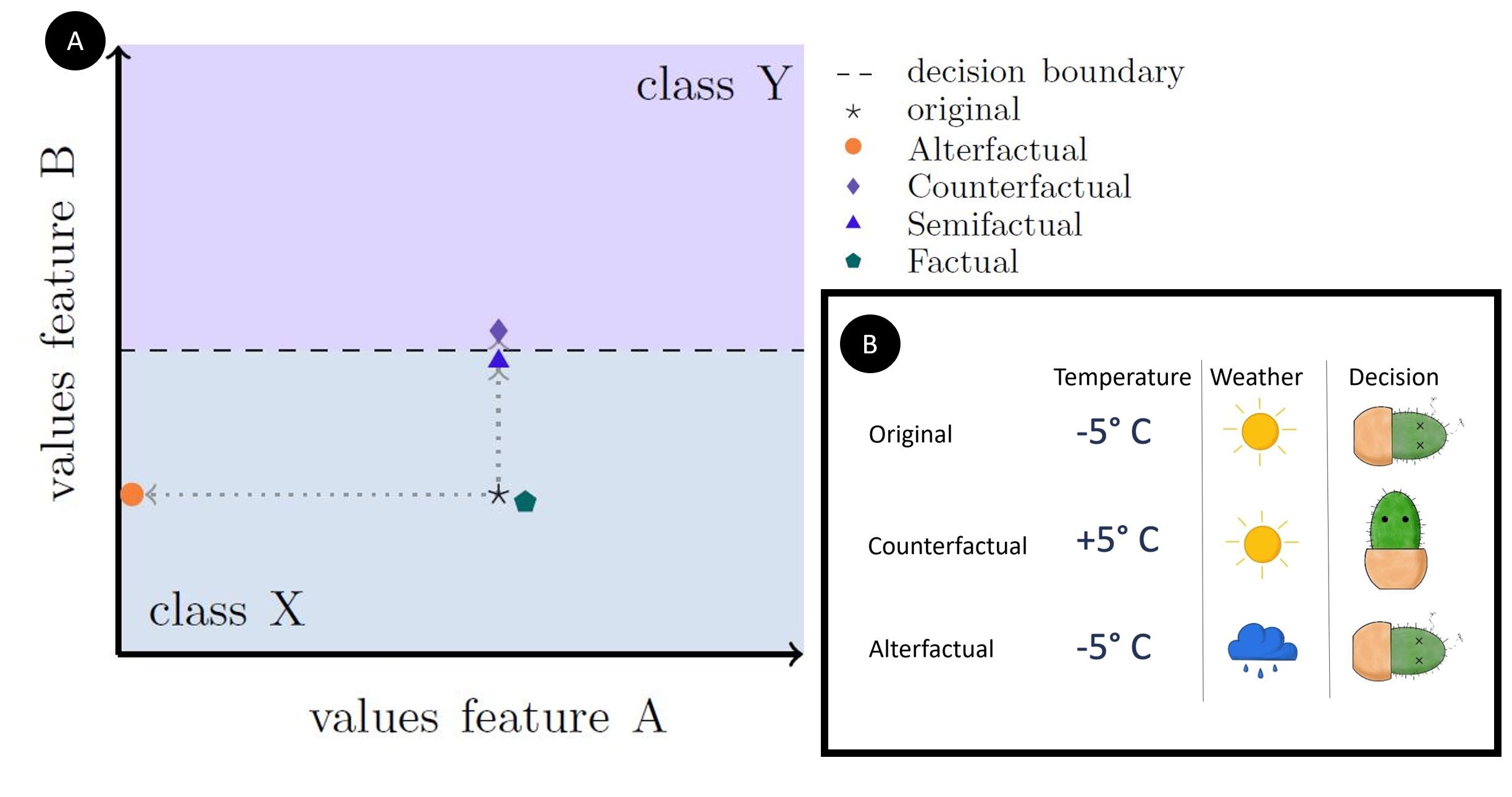}
 \caption{(A) Conceptual comparison of factual, counter-, semi-,
and alterfactual explanations. The diagram shows the original input
which is to be explained below the decision boundary belonging to class X. A
factual explanation could be the nearest neighbor, located anywhere around
the original input. A semifactual explanation would be located in minimal distance
directly next to the decision boundary, but still below it. A counterfactual explanation would be above it in the region of class Y, but barely so. An alterfactual explanation
would move in parallel to the decision boundary, indicating which feature
values would not modify the model's decision. Note that this diagram is
highly simplified - normally, there are more than two features, the decision boundary is more complex, etc.
(B) Examples of a counterfactual and an alterfactual explanation. Input features to a fictional decision system to be explained are \emph{temperature} and \emph{weather}, whereas the former is relevant and the latter is irrelevant to the AI's decision on whether a cactus survives or not.} \label{fig:decision_boundary}
\end{figure}
Prominent examples of these explanatory mechanisms are \emph{Counterfactual Explanations} and \emph{Semifactual Explanations} \citep{kenny2020generating}.
Counterfactual explanations show a version of the input data that is altered just enough to change an AI's decision.  
By doing so, the user is shown not only \emph{which} features are relevant to the decision, but more importantly, \emph{how} they would need to be changed 
to result in a different decision of the AI.
Semifactual explanations follow a similar principle, but they modify the relevant features of the input data to an extent that the AI's decision does not change just yet.

All of these methods have in common that they focus on the \emph{important} features.
However, awareness of irrelevant features can also contribute substantially to the complete understanding of a decision domain, as knowledge of the important features for the AI does not necessarily imply knowledge of the unimportant ones. 

For example, if we want to investigate whether an AI system is subject to some bias regarding its predictions, we often want to know explicitly whether a particular feature is completely irrelevant to a classifier.
As a concrete example, consider an AI system that assesses a person's creditworthiness based on various characteristics, and we want to study that system regarding its fairness. If that system was completely fair, a counterfactual explanation would be of the form: \emph{If your income was higher, you would be creditworthy.}
However, this explanation does not exclude the possibility that your skin color also influenced the AI's decision.
It only shows that the income had a high impact on the AI. 
An explanation confined to the irrelevant features, on the other hand, might say \emph{No matter what your skin color is, the decision would not change.}
In this case, direct communication of irrelevant features ascertains, that the system is fair with regards to skin color.
Conventional counterfactual thinking explanation paradigms do not provide this information directly.

To address this issue, this paper introduces and evaluates a novel explanatory paradigm.
We call explanations that follow this paradigm \emph{Alterfactual Explanations}. 
This principle is based on showing the user of the XAI system an alternative reality that leads to the exact same decision of the AI, but where only irrelevant features change. All relevant features of the input data, on the other hand, remain the same. As this type of explanation conveys completely different information than common methods, we investigate whether the mental model that users have of the explained AI system is also formed in a different way, or can even be improved. 
We show that the communication of features unimportant to the decision contributes significantly to the understanding and formation of a mental model of AI systems.

For this purpose, Section 2 gives an overview of related methods and approaches. In Section 3, the principle of alterfactual explanations is explained in further detail.
In Section 4 we lay the foundation for further research regarding alterfactual explanations by evaluating the potential of our approach in a user study. In this study, we used an imaginary AI to compare how the most widely used counterfactual thinking paradigm, namely counterfactual explanations, compares to our new concept of alterfactual explanations in terms of mental model creation and explanation satisfaction. 
As we regard our approach not as a substitution, but rather as supplement to counterfactual explanations, we further explore the combination of both explanation methods in our user study. Section 5 reports the results of our study. We discuss our findings in Section 6, before we conclude our work and give an outlook into future research in Section 7.

\section{Related Work}
As the approach presented in this paper can be counted to the class of XAI methods that work by inducing counterfactual thinking processes, it is important to gain an understanding of how common methods from this field work. Therefore, this section gives an overview on related concepts that also try to answer the question: \emph{What if...?}

\subsubsection{Factual Explanations}
Factual explanations are the traditional way of explaining by example, and often provide a similar instance from the underlying data set (adapted or not) for the input data point that is to be explained \citep{keane2021twin}. Other approaches do not choose an instance from the dataset, but generate new ones \citep{guidotti2019factual}. The idea behind factual explanations is that similar data instances lead to similar decisions, and the awareness of those similarities leads to a better understanding of the model. Thus, they aim to answer the question \emph{If the input would look more like this Factual Explanation, what would the decision be?}.

\subsubsection{Contrastive Explanations}
Contrastive explanations in the context of XAI answer the question of why a decision occurred relative to some other decision that could also have happened for a specific input \citep{stepin2021survey}. The decision that happened is often called \emph{fact}, while \emph{foil} refers to another possible decision it is being contrasted with \citep{miller2019explanation}. 
Contrastive explanations are usually structured as \emph{Decision X as compared to decision Y occurred because features $f_1$ ... $f_n$ are present and features $f_1^\prime$ ... $f_n^\prime$ are absent} \citep{verma2020counterfactual}. 
They, therefore, highlight the required minimally present and absent features to achieve the given decision.

\subsubsection{Counterfactual Explanations}
Counterfactual explanations are often conflated with contrastive explanations, but do actually state which changes to a given input would be necessary to achieve a contrastive output, i.e.  the foil, by providing an example \citep{stepin2021survey}. Counterfactual explanations are a common method humans naturally use when attempting to explain something and answer the question of \emph{Why not ...?} \citep{miller2019explanation, byrne2019counterfactuals}. In XAI, counterfactual explanations  are  usually  used  for  classification tasks \citep{verma2020counterfactual}. Counterfactual explanations should be minimal, which means they should change as little in the original input as possible to cross the decision boundary of the model between the fact and foil class \citep{keane2021twin, miller2021contrastive}. In the context of generating counterfactuals explanations, the foil class will be referred to as target class. Since counterfactual explanations are a way for a user to understand what they would need to change in order to achieve a more favorable outcome for themselves, many  researchers have emphasized that counterfactual explanations should be actionable and feasible, i.e., should provide a user with an example that is achievable and realistic in real life \citep{barocas2020hidden, ustun2019actionable}. 
In certain scenarios, modern approaches for generating counterfactual explanations have shown significant advantages over feature attribution mechanisms in terms of mental model creation and explanation satisfaction \citep{mertes2020ganterfactual}. 
\citet{wachter2017counterfactual} name multiple advantages of counterfactual explanations, such as being able to detect biases in a model, providing insight without attempting to explain the complicated inner state of the model, and often being efficient to compute. 

\subsubsection{Semifactual Explanations}
Similar to counterfactual explanations, semifactual explanations are an explanation type humans commonly use. They follow the pattern of \emph{Even if X, still P.}, which means that even if the input was changed in a certain way, the prediction of the model would still not change to the foil \citep{mccloy2002semifactual}. In an XAI context, this means that an example, based on the original input, is provided that modifies the input in such a way that moves it toward the decision boundary of the model, but stops just before crossing it \citep{kenny2020generating}. Similar to counterfactual explanations, semifactual explanations can be used to guide a user’s future action, possibly in a way to deter them from moving toward the decision boundary \citep{keane2021twin}. 

\section{The Concept of Alterfactual Explanations}
The basic idea of alterfactual explanations introduced in this paper is to strengthen the user's mental model of an AI by showing irrelevant attributes of a predicted instance. 
Hereby, we understand irrelevance as the property that the corresponding feature, regardless of its value, does not contribute in any way to the decision of the AI model.
When looking at models that are making decisions by mapping some sort of input data $x \in X$ to output data $y \in Y$, the so-called \emph{decision boundary} describes the region in $X$ which contains data points where the corresponding $y$ that is calculated by the model is ambiguous, i.e., lies just between different instances of $Y$. Thus, irrelevant features can be thought of as features that do not contribute to a data point's distance to the decision boundary. 

On the other hand, the information that is carried out by an explanation should be communicated as clearly as possible. As the information that is contained in an alterfactual explanation consists of the \emph{irrelevance} of certain features, it should somehow be emphasized that these features can take \emph{any} possible value. If we would change the respective features only to a small amount, the irrelevance is not clearly demonstrated to the user.
Therefore, we argue that an alterfactual explanation should change the affected features to the maximum amount possible. By doing so, we communicate that the feature, \emph{even if it is changed as much as it can change}, still does not influence the decision.

We take those two considerations as the base for the definition of an alterfactual explanation:


\begin{quote}
Let $d: X \times X \rightarrow \mathbb{R}$ be a distance metric on the input space $X$. An \emph{alterfactual explanation} for a model $M$ is an altered version $a \in X$ of an original input data point $x \in X$, that maximizes the distance $d(x,a)$ whereas the distance to the decision boundary $B \subset X$ and the prediction of the model do not change: $d(x,B)=d(a,B)$ and $M(x)=M(a)$
\end{quote}

Thus, the main difference between an alterfactual explanation and a counterfactual or semifactual explanation becomes clear: While the latter methods alter features resulting in a decreased distance to the decision boundary, the former method tries to keep that distance fixed. Further, while counterfactual explanations as well as semifactual explanations try to keep the overall change to the original input minimal \citep{keane2021if, kenny2020generating}, alterfactual explanations do exactly the opposite, which is depicted in Figure \ref{fig:decision_boundary}A.
Figure \ref{fig:decision_boundary}B illustrates the difference between counterfactual and alterfactual explanations using a simple example.

\section{User Study}
In order to validate if our approach of focusing only on irrelevant features for explaining an AI system helps users to form correct mental models of the system, we performed an online user study. Prior to the real study, a pilot study (n=14) was conducted to find out whether subjects could cope with the tasks.
\subsection{Hypotheses}
The hypotheses that we addressed in our user study are as follows:

\begin{enumerate}
    \item Mental Model Creation
    \begin{enumerate}
        \item Alterfactual explanations lead to a more correct mental model of the AI than no explanations.
        \item Alterfactual explanations lead to similarly good mental models as counterfactual explanations.
        \item The combination of alterfactual and counterfactual explanations outperform both alterfactual as well as counterfactual explanations in terms of mental model creation.
    \end{enumerate}
    \item Explanation Satisfaction
    \begin{enumerate}
        \item Alterfactual explanations lead to a similarly good explanation satisfaction as counterfactual explanations.
        \item The combination of alterfactual and counterfactual explanations outperform both alterfactual as well as counterfactual explanations in terms of explanation satisfaction.
    \end{enumerate}
\end{enumerate}

\subsection{Methodology}
In order to test the hypotheses stated above, an online user study was conducted. We used a between-subject design with four conditions: 
\begin{itemize}
    \item \textbf{Alterfactual condition.} Participants in that condition were presented with original input features to an AI as well as alterfactual explanations.
    \item \textbf{Counterfactual condition.} Participants in that condition were presented with the original features as well as the counterfactual explanations.
    \item \textbf{Combination condition.} Participants in that condition were presented with the original features as well as both the alterfactual and the counterfactual explanations.
    \item \textbf{No Explanation condition.} Participants in that condition were presented only with the original features. No explanation was shown.
\end{itemize}

Between-subject was chosen, mainly because we wanted to avoid order effects and mitigate the risk of fatigue. 
In the study, the participants were presented with an imaginary AI.
The participants were told that the AI decides if hypothetical historical documents are forged or not. This specific scenario was chosen as it is not present in most people's everyday life, ensuring that the mental model of the AI that the participants develop is predominantly induced by the explanations that they are presented with during the study and do not stem from prior knowledge of the domain. 
The AI gets different inputs to work with. We designed the imaginary AI so that it follows a set of rules (unknown to the participants), where each input feature has a specific relevance to the AI. Those features are as follows:
\begin{itemize}
    \item \textbf{Parchment Color.} The documents can be either of \emph{light}, \emph{medium} or \emph{dark} parchment.
    \item \textbf{Word Count.} A single integer in the range $[1,500]$.
    \item \textbf{Year of Creation.} The documents were created sometime between 200 BC and 200 AD.
\end{itemize}
The rules which the fictional AI uses to decide if a document is forged are:
\begin{itemize}
    \item A document is forged if the word count is equal to or below 50.
    \item A document is forged if the word count is between 51 and 150 \emph{and} the parchment color is light or medium.
    \item In all other cases, the document is considered to be authentic
\end{itemize}

Therefore, one attribute is always relevant (word count), one is relevant only in some cases (parchment color), and one is always irrelevant (year of creation).
After answering some questions about their demographic background, the participants were given some general information about the data and AI used in the experiment. They were told that some historical documents had been found, and some of them had already been identified as forgeries.
Futhermore, they were told that an AI had been trained to detect forgeries based on a short description of the documents containing the three attributes mentioned above. The three attributes were shown along with their value ranges. An exemplary input to the fictional AI was displayed in a table. Additionally, we explained which explanation type the participant was going to be shown during the study, and how the explanation type works. The participants were provided with example explanations that could be revealed by clicking a button. After using that button, the explanations were shown next to the original input. An example explanation is shown in Figure \ref{fig:descriptor}. 
Following this introduction, the participants were given two example inputs and corresponding explanations in order to familiarise themselves with the document descriptors and the mechanism to reveal the explanations. After that, each participant was quizzed about the information that was given up to that point. By doing so, we could exclude subjects who did not conscientiously participate in the study.
After the quiz, a short training phase followed. In this phase, the participants were shown four exemplary document descriptors. Explanations for the AI's decisions, as well as the decisions itself were shown as well. The training phase was conducted to give the participants another chance to get comfortable with the explanation type and the domain itself. Subsequently, the study itself started. It was divided into three parts: For assessing the participants' mental model of the AI, we used \emph{(i)} a prediction task and \emph{(ii)} a questionnaire about the AI's rule set. To assess the participants' explanation satisfaction, we used \emph{(iii)} an explanation satisfaction questionnaire.

\subsubsection{Mental Model Creation (i): Prediction Task}
The goal of the prediction task was to detect how well the participants could anticipate the classifier's decisions, which provides a quick window into how well they \emph{understood} the AI \citep{hoffman2018metrics}. 
To this end, eight example inputs with explanations were shown in a random order. Four examples were classified as \emph{forged} by the AI, whereas four examples were predicted as being \emph{authentic}. As proposed by \citet{hoffman2018metrics}, the decision of the AI was \emph{not} shown, but had to be predicted by the participants. The idea of such a prediction task is that a good explanation should help to build a correct mental model of the AI, allowing to understand its decision process to an extent that those decisions can be predicted by the user. Additionally to the prediction of the AI's decision, participants had to choose how confident they were in their prediction on a 7-point Likert scale (0~=~not at all confident, 6~=~very confident). Furthermore, they had to justify their prediction in a free text form.
Participants that were in the \emph{No Explanation} condition did not see any explanations but had to rely on the original input data for their predictions.
For every single prediction task, explanations had to be revealed by pressing the \emph{Explain} button. By doing so, we were able to track if the participants really looked at the explanations. 

\subsubsection{Mental Model Creation (ii): Understanding Questionnaire}
To assess if the participants developed a correct mental model of the AI's decision process, for each feature (i.e., parchment color, word count, year of creation), they were explicitly asked how much they agreed that it was relevant to the AI's decision on a 5-point Likert scale (0~=~strongly disagree, 4~=~strongly agree) after completing all predictions of the Prediction Task. 
Thus, while the Prediction Task can be seen as implicit measurement of the mental model's correctness, our Understanding Questionnaire directly measures if participants understood the relevance of different features.

\subsubsection{Explanation Satisfaction}
In order to validate hypotheses 2a and 2b, we used the Explanation Satisfaction Scale proposed by \citet{hoffman2018metrics} which consists of seven items, rated on a 5-point Likert scale (0~=~strongly disagree, 5~=~strongly agree).

Finally, the participants had the possibility to give free text feedback. The whole study was built using the \emph{oTree} framework by \citet{chen2016otree}.

\subsection{Participants}

113 Participants between 24 and 71 years (\textit{M}~=~41.2, \textit{SD}~=~10.2) were recruited via Amazon MTurk. 62 of them were male, 48 female, 1 non-binary, and 2 preferred not to answer this question. Only participants with an \emph{MTurk Masters Qualification} were allowed to participate, and subjects that did not pass the quiz were excluded from the study to minimize bias due to unconscious participants. The participants were randomly separated in the four conditions. Subjects of the three explanation conditions that did not look at a single explanation during the whole study were moved to the \emph{No Explanation} condition for evaluation. Participants got paid a base reward of 5.00\$ and another 0.50\$ for each right prediction in the Prediction Task. By communicating that bonus payment before participation, we wanted to further motivate the participants to stay focused on the study.
Only 5.3\% of the participants had no experience with AI. Most of the participants (86.7\%) have heard from AI in the media. In general, 79.7\% of the participants were expecting a positive or extremely positive impact of AI systems in the future. 

\section{Results}

\subsection{Mental Model Creation}
To investigate the impact of the four different experimental conditions\footnote{(\emph{Alterfactual} condition, \emph{Counterfactual} condition, \emph{Combination} condition, \emph{No Explanation} condition)} on the (1) understanding and (2) prediction accuracy, we conducted a MANOVA. We found a significant difference, Pillai's Trace~=~0.13, \textit{F}(6,218)~=~2.52, \textit{p}~=~.022. 

The following ANOVA revealed that only the understanding of the participants showed significant differences between the conditions:
\begin{itemize}
    \item \emph{Understanding}: \textit{F}(3,109)~=~3.90, \textit{p}~=~.011.
    \item \emph{Prediction Accuracy}: \textit{F}(3,110)~=~2.63, \textit{p}~=~.217.
\end{itemize}

\begin{figure*}
    \centering
    \captionsetup{justification=raggedright,margin=.5cm}
     \begin{minipage}{0.44\textwidth}
 \includegraphics[width=1\linewidth]{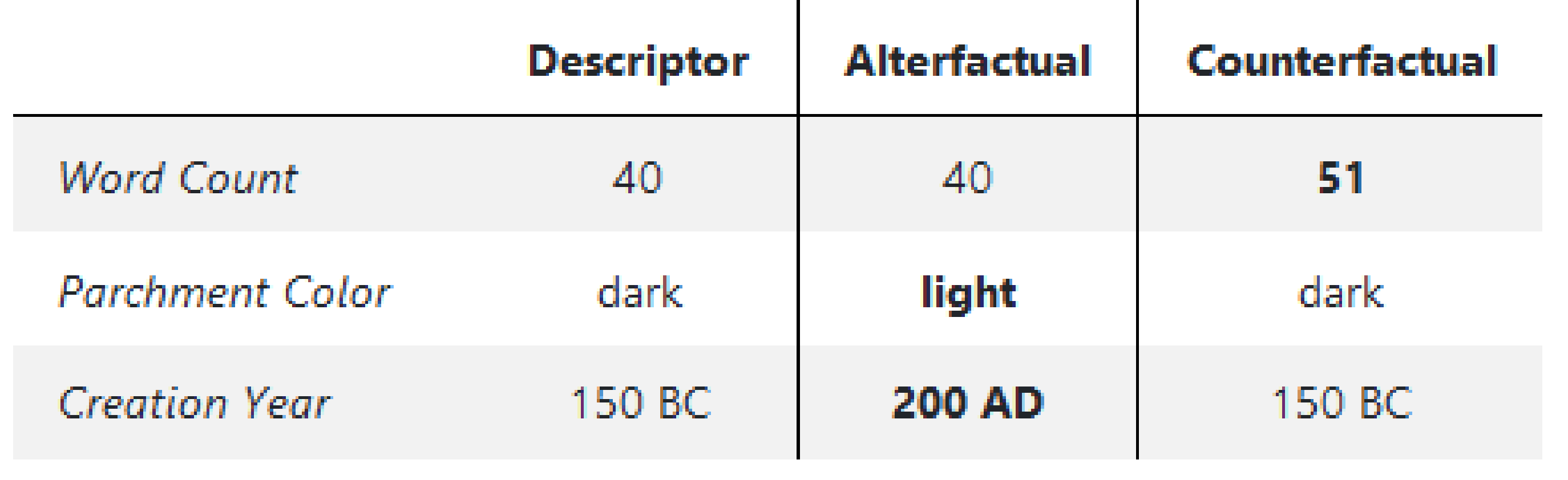}
 \caption{A sample document descriptor with  explanations. In the \emph{Combination} condition, both an alter- and a counterfactual explanation were shown. Subjects in the \emph{Alterfactual} and \emph{Counterfactual} conditions did not see the respective other explanation type. Subjects in the \emph{No Explanation} condition did not see an explanation at all, but only the original document descriptor.} \label{fig:descriptor}
\end{minipage}
    \begin{minipage}{0.55\textwidth}
        \includegraphics[width=.5\linewidth]{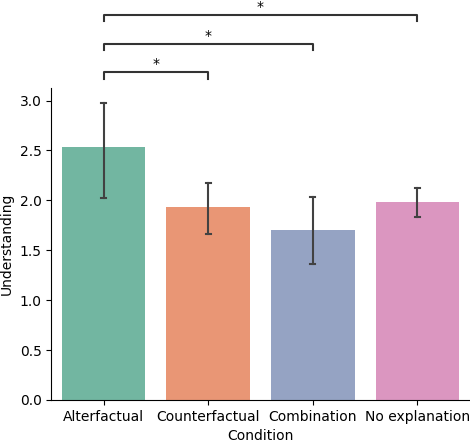}
        \includegraphics[width=.5\linewidth]{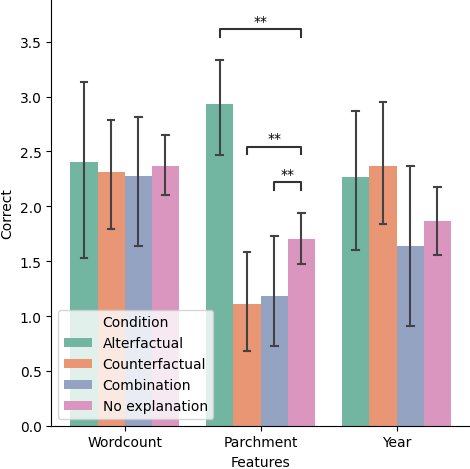}
         \caption{Impact of the four experimental conditions on the understanding of the relevant features of the AI. Alterfactual explanations outperformed all other conditions in helping participants to understand the relevant features of the AI system. Best viewed in color. Error bars represent the 95\% CI. *\textit{p}~$<$~.05,  **\textit{p}~$<$~.001.}
    \label{fig:MentalModelCreation}
    \end{minipage}
    \vspace{-0.3cm}
\end{figure*}

As displayed in Figure \ref{fig:MentalModelCreation}, the post-hoc t-tests showed that the participants' understanding was significantly better in the \emph{Alterfactual} condition compared to all other conditions. The effect size \textit{d} is calculated according to \citet{cohen2013statistical}\footnote{Interpretation of the effect size is: \textit{d}~$<$~.5~:~small effect; \textit{d}~=~0.5-0.8~:~medium~effect; \textit{d}~$>$~0.8~:~large~effect}:

\begin{itemize}
    \item \textbf{alterfactual vs. counterfactual}: \textit{t}(109)~=~2.58, \textit{p}~=~.011, \textit{d}~=~0.89 (large effect).
    \item \textbf{alterfactual vs. combination}: \textit{t}(109)~=~3.11, \textit{p}~=~.002, \textit{d}~=~1.24 (large effect).
    \item \textbf{alterfactual vs. no explanation}: \textit{t}(109)~=~2.86, \textit{p}~=~.005, \textit{d}~=~0.82 (large effect).
\end{itemize}

The results indicate that alterfactual explanations help participants understand the relevant features more correctly than in all other conditions. Interestingly, the combination of alterfactual and counterfactual explanations leads to a worse performance and understanding by the participants (see Figure \ref{fig:MentalModelCreation}).

Therefore, hypothesis 1a holds, because alterfactual explanations outperformed the \emph{No Explanation} condition as well as the \emph{Combination} condition. Hypotheses 1b and 1c have to be rejected because alterfactuals explanations also outperfomed counterfactual explanations as well as the combination of both explanation types in the context of mental model creation.  

Wondering about the results, especially about the fact that the \emph{No Explanation} condition outperformed the \emph{Combination} condition, we took a closer look, which of the features (i.e., word count, parchment color, year of creation) the participants did or did not understand in each condition. For this, we compared the amount of the correct features between the group, using a MANOVA.
We found a significant difference, Pillai's Trace~=~0.26, \textit{F}(9,327)~=~3.49, \textit{p}~$<$~.001. 

The following ANOVA revealed that only the feature \emph{parchment color} showed significant differences between the conditions:
\begin{itemize}
    \item \emph{Parchment color}: \textit{F}(3,109)~=~10.49, \textit{p}~$<$~.001.
    \item \emph{Word count}: \textit{F}(3,109)~=~0.03, \textit{p}~=~.099.
    \item \emph{Creation year}: \textit{F}(3,109)~=~1.22, \textit{p}~=~.305.
\end{itemize}

As displayed in Figure \ref{fig:MentalModelCreation}, the post-hoc t-tests showed that the participants' correct understanding of the relevance of the parchment color feature were significant better in the \emph{Alterfactual} condition, compared to all other conditions:

\begin{itemize}
    \item \textbf{alterfactual vs. counterfactual}: \textit{t}(109)~=~5.21, \textit{p}~$<$~.001, \textit{d}~=~1.80 (large effect).
    \item \textbf{alterfactual vs. combination}: \textit{t}(109)~=~4.34, \textit{p}~$<$~.001, \textit{d}~=~1.72 (large effect).
    \item \textbf{alterfactual vs. no explanation}: \textit{t}(109)~=~4.24, \textit{p}~$<$~.001, \textit{d}~=~1.21 (large effect).
\end{itemize}

\subsection{Explanation Satisfaction}
The ANOVA revealed that there were no significant differences between the three explanation conditions, \textit{F}(2,42)~=~1.57, \textit{p}~=~.219, indicating that participants felt not specific satisfied by one of the explanation conditions. 

Therefore, hypothesis 2b has to be rejected, as the combination of alterfactual and counterfactual explanations does not lead to a higher explanation satisfaction of the participants. Nevertheless, hypothesis 2a holds since the alterfactual explanations do not differ significantly compared to counterfactual explanations. 

\section{Discussion}
The results of our user study show novel insights into the explanatory performance of the different XAI approaches. \newline
\textbf{Alterfactual Explanations Support Global Understanding of Users.}
First of all, although not significantly differing from the other conditions in the Prediction Task, subjects that were provided with alterfactual explanations performed significantly better in the Understanding Task than all other participants. This indicates that direct communication of information about irrelevant features does indeed offer benefits. Contrary to our original assumption, the alterfactual explanations outperformed even the more traditional counterfactual explanations. Different from the Prediction Task, the Understanding Task directly surveys the users' mental models regarding the relevance of the input features. Thus, we argue that alterfactual explanations work better when it comes to the communication of how important different features are for a decision in general, although they do not convey a better understanding of which exact decision will be made when presented with a concrete input sample compared to counterfactual explanations. This suggests that alterfactual explanation could find application in scenarios where a global understanding of the AI system is important. Our investigation of the participants' feature-specific understanding strengthens this assumption: The alterfactual explanations' better performance in the Understanding Task mainly stems from users presented with alterfactual explanations having a significantly better understanding of the importance of the \emph{parchment} feature. As that feature was relevant in some cases and irrelevant in others, understanding its relevance highly depends on global understanding of the model.
However, future studies have to be conducted to assess the capability of alterfactual explanations to induce a global understanding of an AI's decision process in a broader scope.\newline
\textbf{Too many Explanations can Overstrain Users.}
It seems very surprising that the combination of alterfactual and counterfactual explanations performs poorly, although they contain more information than any other condition. We assume that this stems from the fact that more information comes with higher demands on the users' attentiveness. We argue that the participants in the \emph{Combination} condition were simply overwhelmed by the wealth of information. This finding is in line with \emph{cognitive load} research that emphasized the fact that too much information can overwhelm users \citep{sweller1998cognitive}. Future research has to find ways to communicate this vast amount of information without overburdening users.\newline
\textbf{Alterfactual Explanations are Equally Satisfying as Counterfactual Explanations.}
Furthermore, we found no significant differences regarding Explanation Satisfaction between the three conditions that were presented with some kind of explanation. We argue again that the combination of counterfactual and alterfactual explanations could have overwhelmed the user also in this regards. However, we see that alterfactual explanations lead to similarly good Explanation Satisfaction as the traditional counterfactual explanations, making them a viable approach for real-world XAI scenarios.

\section{Conclusion \& Outlook}
In this work, we presented a new XAI paradigm that we call \emph{Alterfactual Explanations}. Our approach is based on only communicating information about features that are irrelevant to an AI's decision. A user study that we conducted showed that alterfactual explanations show huge potential for the field of XAI. In an Understanding Task measuring the capabilities of users to tell which features of an example input to an AI are important for its decision, alterfactual explanations significantly outperformed the more traditional counterfactual explanations as well as the combination of alterfactual and counterfactual explanations. Surprisingly, combining counterfactual and alterfactual explanations did not result in more correct mental models. We showed that alterfactual explanations lead to a similar good Explanation Satisfaction as counterfactual explanations.
As alterfactual and counterfactual explanations convey a different kind of information, future research has to investigate how the combination of the two can leverage the best of both worlds to create even better explanations without overwhelming the user. 
Further, our study did not include a comparison to Semifactual Explanations. As the concept of those is also based on showing an alternative reality in which the decision does not change, it is likely that users get confused by the differences between Alterfactual and Semifactual explanations. Therefore, the advantages and disadvantages of those two concepts have to be evaluated in further research. 
\bibliographystyle{named}
\bibliography{main}
\end{document}